\pdfoutput=1

\documentclass[11pt]{article}

\usepackage[preprint]{coling}

\usepackage{times}
\usepackage{latexsym}

\usepackage[T1]{fontenc}

\usepackage[utf8]{inputenc}

\usepackage{microtype}

\usepackage{inconsolata}

\usepackage{graphicx}

\usepackage{amsmath}
\usepackage{booktabs}
\usepackage{float}
\usepackage{xcolor}
\usepackage[most]{tcolorbox}

\title{SelfPrompt: Autonomously Evaluating LLM Robustness via\\Domain-Constrained Knowledge Guidelines and\\ Refined Adversarial Prompts}

\author{Aihua Pei \\Waseda University\\
  aika@fuji.waseda.jp\And
    Zehua Yang \\ Waseda University \\ yangzehua@akane.waseda.jp\And
    Shunan Zhu \\ Waseda University \\ shunan-zhu@ruri.waseda.jp
    \AND
    Ruoxi Cheng \\ Southeast University \\ 213200761@seu.edu.cn \And
    Ju Jia \\ Southeast University \\ jiaju@seu.edu.cn
}

\begin{document}
\maketitle
\begin{abstract}
Traditional methods for evaluating the robustness of large language models (LLMs) often rely on standardized benchmarks, which can escalate costs and limit evaluations across varied domains. This paper introduces a novel framework designed to autonomously evaluate the robustness of LLMs by incorporating refined adversarial prompts and domain-constrained knowledge guidelines in the form of knowledge graphs. Our method systematically generates descriptive sentences from domain-constrained knowledge graph triplets to formulate adversarial prompts, enhancing the relevance and challenge of the evaluation. These prompts, generated by the LLM itself and tailored to evaluate its own robustness, undergo a rigorous filtering and refinement process, ensuring that only those with high textual fluency and semantic fidelity are used. This self-evaluation mechanism allows the LLM to evaluate its robustness without the need for external benchmarks. We assess the effectiveness of our framework through extensive testing on both proprietary models like ChatGPT \cite{OpenAI2024ChatGPT} and open-source models such as Llama-3.1 \cite{llama2024}, Phi-3 \cite{phi32024}, and Mistral \cite{mistral2024}. Results confirm that our approach not only reduces dependency on conventional data but also provides a targeted and efficient means of evaluating LLM robustness in constrained domains.
\end{abstract}

\section{Introduction}


Large language models (LLMs) have garnered significant attention due to their exceptional performance across various natural language processing (NLP) tasks. However, as these models are widely applied in critical domains, they also face the risk of adversarial attacks triggered by prompts. Adversarial attacks aim to mislead models into making incorrect judgments through carefully designed prompts, potentially causing severe damage to users. Therefore, it is necessary to assess the robustness of models against adversarial attacks using robustness evaluations.

Existing adversarial robustness evaluation frameworks for large language models (LLMs), like AdvGLUE \cite{wang2021adversarial} and PromptAttack \cite{xu2023llm}, use specialized benchmark datasets that require extensive manual annotation. This not only limits their applicability but also increases operational costs. Moreover, when LLMs are used in constrained domains such as medicine or biology, the mismatch between generic benchmark datasets and the constrained context can lead to inaccurate robustness evaluations. These limitations decrease practicality of the frameworks and complicate the robustness evaluation of LLMs.

This paper proposes an adversarial attack framework (SelfPrompt) that requires the evaluated LLMs themselves to utilize domain-constrained knowledge guidelines to generate and poison prompts from knowledge graph triplets, thereby assessing their robustness. The generation of adversarial prompts is meticulously refined to optimize quality and evaluation effectiveness, while ensuring that the quality of adversarial prompts generated by different large language models is relatively consistent. We apply this framework to generate prompts from both general and constrained domain knowledge graphs, evaluating the resilience of multiple LLMs under adversarial attack conditions. Specifically, our contributions include:

\begin{itemize}
\item This paper introduces a framework that allows large language models (LLMs) to autonomously evaluate their robustness in constrained domains by generating adversarial prompts from domain-specific knowledge graph triplets. This method enhances the practical relevance of robustness evaluations by tailoring the prompts to the specific operational domains of the LLMs.

\item To ensure stable quality of adversarial prompts across various large models and maintain comparability in their robustness evaluations, we employ a filter. This filter assesses the text fluency and semantic fidelity of the prompts, allowing us to refine and exclude those that do not meet our quality criteria.

\item We confirm that the robustness of large language models is influenced by the domain of knowledge corresponding to the prompts. The robustness of the same large language model measured on general or constrained domain knowledge graphs is not similar. While models with larger parameters in the same series tend to exhibit stronger robustness in general domains, this is not necessarily the case in constrained domains. Therefore, it is crucial to consider the differences in knowledge domains when evaluating robustness of LLMs.
\end{itemize}

\section{Related Works}
\subsection{Robustness Evaluation of LLMs}
Large language models (LLMs), such as the ChatGPT family and the Llama family, have attracted much attention for their excellent performance in a variety of natural language processing tasks \cite{touvron2023llama,brown2020language}. However, as these models are widely used in critical domains applications, evaluating their robustness has also become a hot research topic. There are four main streams of work \cite{li2023survey, ailem2024examining, zhuo2023robustness} on robustness research: robustness under distribution shift \cite{yang2023glue}, robustness to adversarial attacks \cite{wang2023robustness, zhu2023promptbench}, robustness to prompt formats and instruction templates \cite{mizrahi2023state,voronov2024mind,weber2023mind} and robustness to
dataset bias \cite{gururangan2018annotation, niven2019probing, le2020adversarial}. Our work focus on evaluating robustness to adversarial attacks of LLMs.

Adversarial attacks aim to mislead the model to make wrong judgments through well-designed inputs, while adversarial robustness evaluation attempts to determine and enhance robustness of the model to these attacks. Current robustness evaluation frameworks for LLMs are mainly based on specially constructed benchmark datasets (e.g., the GLUE dataset \cite{wang2018glue} and ANLI dataset \cite{nie2020adversarial}) for evaluating natural language comprehension capabilities of LLMs \cite{goel2021robustness}.

 AdvGLUE \cite{wang2021adversarial} and AdvGLUE++ \cite{wang2023decodingtrust} are two frameworks specifically designed to evaluate the adversarial robustness of language models. These frameworks challenge the ability to make judgments under complex and subtle semantic changes by providing a series of adversarial samples of models. AdvGLUE++ is a further extension of AdvGLUE that introduces more adversarial samples, especially for new emerging LLMs such as the Alpaca and Vicuna families \cite{taori2023alpaca, vicuna2023}.PromptAttack enhance the attack power by ensembling adversarial examples at different perturbation levels \cite{xu2023llm}. These evaluation frameworks exhibit a common feature: testing and improving the robustness of the model by constructing inputs that may cause the model to misjudge. These inputs include both subtle textual modifications and complex semantic transformations, aiming to comprehensively evaluate robustness of the model to various challenges that may be encountered in real-world applications. 

\subsection{Adversarial Prompt Generation from Knowledge Graphs}

In evaluating robustness of LLMs, we need to know whether they have such knowledge and whether they can accurately express their knowledge. Knowledge graphs can help us generate adversarial attack prompt with different diversities and complexities. Knowledge graph (KG) is a graph structure for representing knowledge, where nodes represent entities or concepts and edges represent relationships between these entities or concepts.

Some works use different methods to utilize triplet from knowledge graphs generating questions \cite{seyler2017knowledge, kumar2019difficulty, chen2023toward}. Some works utilize the ability of LLMs to generate questions from KGs \cite{guo2022dsm,axelsson2023using}. Recent works \cite{luo2023systematic, luo2024biaskg} also discussed on evaluating factual knowledge of LLMs with the diverse and well-coverage questions generated from KGs and how KGs can be used to induce bias in LLMs.

\subsection{Few-Shot Strategy}
As the popularity of machine learning models, especially large language models (LLMs), continues to grow, the few-shot learning strategy has also garnered significant attention \cite{logan2021cutting, meng2024advancing}. Few-shot learning involves training models with a limited number of samples to perform well on various tasks, minimizing the need for large annotated datasets. This approach is particularly valuable in situations where obtaining extensive labeled data is challenging or costly. By leveraging pre-trained models, few-shot learning allows LLMs to generalize effectively from just a few examples, making it a powerful tool for tasks like text classification, translation, and summarization.

The few-shot learning strategy is designed to enhance model performance in data-scarce environments, which is crucial for applying LLMs to specialized domains where data is often limited. The core of this strategy lies in its ability to utilize prior knowledge embedded in pre-trained models, enabling them to adapt to new tasks quickly and efficiently with minimal data. This adaptability helps uncover the potential of LLMs in diverse applications while maintaining robustness and relevance in domain-specific contexts.

\section{Methodology}

In this section, we first illustrate a robustness self-evaluation framework for large language models based on domain-constrained knowledge guidelines, utilizing adversarial prompt attacks, which we call SelfPrompt. Then, we employ a filter module to ensure the text fluency and semantic fidelity of the adversarial prompts generated by SelfPrompt. Finally, we introduce the metrics for evaluating the robustness of LLMs. All the prompt templates mentioned in this section can be found in the Appendix C.

\subsection{Framework of SelfPrompt} 

Initially, we process the triplets of the knowledge graph to assign them distinct labels. Subsequently, we transform these triplets into original prompts. Finally, these original prompts are converted into adversarial prompts. Next, we provide a detailed description of each step in this process.

\textbf{Labeling Knowledge Graph Triples.} We let $\mathcal{D} = \left\{ (s_i, p_i, o_i) \right\}_{i=1}^{N}$ be the domain-constrained knowledge graph dataset. For each triplet $t = (s, p, o) \in \mathcal{D}$, $s$ refers to the subject of this triplet, while $p$ and $o$ refer to the predicate and object of the triplet, respectively. For example, for the triple $(\textit{Alan Turing}, \textit{field of work}, \textit{logic})$, it has the subject (\textit{Alan Turing}), the predicate (\textit{field of work}), and the object (\textit{logic}). This triple means that \textit{Alan Turing works in the field of logic}.

Considering the structural characteristics of the triples, each triple is labeled with one of the following three labels: \textit{true}, \textit{entity\_error}, and \textit{predicate\_error}. By default, all triples extracted from the knowledge graph are initially labeled as \textit{true}. For each triplet $t = (s, p, o) \in \mathcal{D}$, its label $l$ is randomly assigned to one of the three labels with equal probability, generating incorrect subject, predicate, and object, denoted as $s'$, $p'$, and $o'$, respectively. The modified triple $t'$ is:

\begin{align}
t' = 
\begin{cases}
(s, p, o), & l = \textit{true} \\
(s, p', o), & l = \textit{predicate\textunderscore error} \\
(s', p, o) \text{ or } (s, p, o'), & l = \textit{entity\textunderscore error}
\end{cases}
\end{align}

For example, for the original triple labeled as \textit{true}, $(\textit{Alan Turing}, \textit{field of work}, \textit{logic})$; if it is to be labeled as \textit{predicate\textunderscore error}, it can be modified to $(\textit{Alan Turing}, \textit{position played on team}, \textit{logic})$, which means \textit{Alan Turing plays in the logic position}; the modified predicate is used to describe the \textit{position or specialism of a player on a team}. If it is to be labeled as \textit{entity\textunderscore error}, the original triple can be modified to $(\textit{Richard Wagner}, \textit{field of work}, \textit{logic})$ or $(\textit{Alan Turing}, \textit{field of work}, \textit{Opera})$. The labeled knowledge graph dataset is $\mathcal{D'} = \left\{ (t'_i, l_i) \right\}_{i=1}^{N}$

\begin{figure}[h]
    \centering
    \includegraphics[width=1 \linewidth]{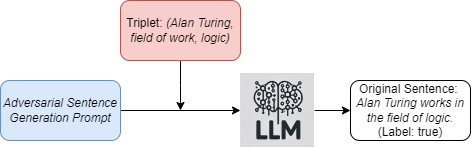}
    \caption{LLM-Based Strategy Example}
    \label{fig:t2p}
\end{figure}

\textbf{Generating Original Prompts.} LLMs are more suitable for handling continuous prompt text rather than structured triplets. For converting triplets into prompts, we offer two strategies: Template-based and LLM-based. The template-based strategy uses templates built into the predicates of the triplets to generate original prompts by replacing these placeholders with specific names. For example, for the triplet $(\textit{Alan Turing}, \textit{field of work}, \textit{logic})$, the template built into the predicate \textit{field of work} is "\textit{[X] works in the field of [Y]}." By replacing \textit{[X]} with \textit{Alan Turing} and \textit{[Y]} with \textit{logic}, the sentence "\textit{Alan Turing works in the field of logic}" is generated. The LLM-based strategy involves feeding triplets to the LLM whose robustness is being evaluated, which then generates descriptive sentences based on these elements. Figure~\ref{fig:t2p} shows the structure of this strategy.

The sentences generated by the two strategies are filled into the corresponding positions of the prompt templates to create original prompts, requiring the LLM to classify these sentences according to the three labels described in this subsection.

\textbf{Constructing Adversarial Prompts.} Adversarial prompts maintain the same main structure as the original prompts and require the LLM, whose robustness is being evaluated, to modify the sentences generated from the triplets to create adversarial sentences. These adversarial sentences should retain the same semantics as the original sentences but lead the LLM to misclassify them. Adversarial prompts are generated by replacing the corresponding parts in the original prompts with adversarial sentences. In the prompt template (\textit{Adversarial Sentences Generation Prompt}) for this step, we provide both the triplets and the sentences generated from them according to the procedure described in this subsection.

\begin{figure}[h]
    \centering
    \includegraphics[width=1 \linewidth]{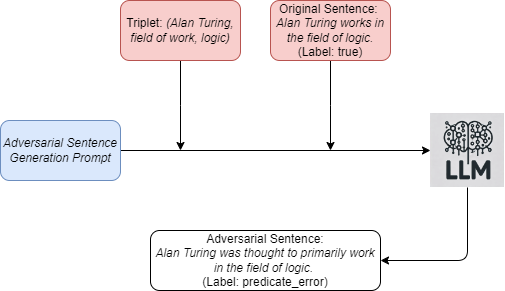}
    \caption{Adversarial Sentence Generation}
    \label{fig:AdversarialPromptGeneration}
\end{figure}

To generate adversarial sentences, we offer an optional few-shot approach that enhances the LLM's ability to produce adversarial sentences by providing example samples that demonstrate the transformation of original sentences into adversarial sentences.

\subsection{Filter Module} 

In the SelfPrompt framework described in section 3.1, we use the LLM being evaluated for robustness to generate both original prompts and adversarial prompts. The quality of adversarial prompts generated by different LLMs varies, posing challenges for the cross-comparison of robustness evaluation results across different LLMs. To address this issue, we designed a filter module to eliminate adversarial sentences that do not meet the criteria for text fluency or semantic fidelity. This ensures that the adversarial prompts generated by different LLMs are of comparable quality, thereby enhancing the reliability and comparability of the robustness evaluation results.For an original sentence $s_{\text{ori}}$ and its corresponding adversarial sentence $s_{\text{adv}}$, the text fluency of $s_{\text{adv}}$ is $tf(s_{\text{adv}})$, the semantic fidelty of $s_{\text{adv}}$ relative to $s_{\text{ori}}$ is $sf(s_{\text{adv}}, s_{\text{ori}})$. Assume that the filtering thresholds for text fluency and semantic fidelity are $\tau_{\text{t}}$ and  $\tau_{\text{s}}$ respectively, the formula for the filter module is as follows:

\begin{align}
\text{f}(s_{\text{adv}}, s_{\text{ori}}) = (\text{tf}(s_{\text{adv}}) > \tau_{\text{t}}) \land (\text{sf}(s_{\text{adv}}, s_{\text{ori}}) > \tau_{\text{s}})
\end{align}

The function \( \text{tf}(s) \) calculates the text fluency of a sentence \( s \) by computing the perplexity of a language model's output for \( s \). The perplexity is defined as:

\begin{align}
P(s) = e^{\text{Loss}(s)}
\end{align}

\noindent{}where \( \text{Loss}(s) \) is the negative log-likelihood loss of predicting the tokens in \( s \) \cite{Goodfellow-et-al-2016}. To manage the typically large values of perplexity, a logarithmic transformation is applied:

\begin{align}
LogP(s) = \log(P(s) + e - 1)
\end{align}

The text fluency score is computed as:

\begin{align}
\text{tf}(s) = \frac{e^{-k / LogP(s)} - 1}{e^{-k} - 1}
\end{align}

\noindent{}where \( k > 0 \); in this experiment, \( k \) is set to 5.

The function \( \text{sf}(s_{\text{adv}}, s_{\text{ori}}) \) computes the semantic fidelity between \( s_{\text{adv}} \) and \( s_{\text{ori}} \) by first calculating the cosine similarity between their embedding vectors \( \textbf{v}_{\text{adv}} \) and \( \textbf{v}_{\text{ori}} \), where:

\begin{align}
\textbf{v}_{\text{adv}} = \text{get\_embedding}(s_{\text{adv}})
\end{align}

\begin{align}
\textbf{v}_{\text{ori}} = \text{get\_embedding}(s_{\text{ori}})
\end{align}

The cosine similarity \cite{Manning:2008:IIR:1394399} is given by:

\begin{align}
\text{cos\_sim}(\textbf{v}_{\text{adv}}, \textbf{v}_{\text{ori}}) = \frac{\textbf{v}_{\text{adv}} \cdot \textbf{v}_{\text{ori}}}{\|\textbf{v}_{\text{adv}}\| \|\textbf{v}_{\text{ori}}\|}
\end{align}

It then scales the cosine similarity to the range [0, 1] using the formula:

\begin{align}
\text{sf}(s_{\text{adv}}, s_{\text{ori}}) = \frac{e^{t \cdot \text{cos\_sim}(\textbf{v}_{\text{adv}}, \textbf{v}_{\text{ori}})} - e^{-t}}{e^t - e^{-t}}
\end{align}

\noindent{}where \( t > 0 \). In this experiment, \( t \) is set to 5.

\subsection{Metrics for Robustness Evaluation}

From a knowledge graph triplet dataset \( \mathcal{D} = \left\{ (s_i, p_i, o_i) \right\}_{i=1}^{N} \), we generate an original prompt set \( \mathcal{O} \) and a corresponding adversarial prompt set \( \mathcal{A} \) of size \( M \) (where \( 0 < M \leq N \), and all elements in set \( \mathcal{A} \) must pass the filter module test described in section 3.2). Let the accuracy of the LLM on the classification task for set \( \mathcal{O} \) be \( \text{ACC}_{\mathcal{O}} \) and for set \( \mathcal{A} \) be \( \text{ACC}_{\mathcal{A}} \)(both \( \text{ACC}_{\mathcal{O}} \) and \( \text{ACC}_{\mathcal{A}} \) range from 0 to 1). The robustness metric \( R(\text{ACC}_{\mathcal{A}}, \text{ACC}_{\mathcal{O}}) \) evaluates a model's ability to handle adversarial prompts. It is defined as:

\begin{equation}
\small
R(\text{ACC}_{\mathcal{A}}, \text{ACC}_{\mathcal{O}}) = \sin\left(\frac{\pi}{2} \cdot \text{ACC}_{\mathcal{A}} \cdot \left(1 - \frac{\text{ACC}_{\mathcal{O}}^j}{j}\right)\right)
\label{eq:robustness_formula}
\end{equation}

where \( \text{ACC}_{\mathcal{A}} \) is positively correlated with robustness since a higher \( \text{ACC}_{\mathcal{A}} \) reflects better resistance to adversarial attacks. Conversely, \( \text{ACC}_{\mathcal{O}} \) is negatively correlated with robustness because, in most cases, \( \text{ACC}_{\mathcal{A}} < \text{ACC}_{\mathcal{O}} \). When \( \text{ACC}_{\mathcal{A}} \) is the same, a lower \( \text{ACC}_{\mathcal{O}} \) indicates that the LLM is less influenced by adversarial attacks, leading to a higher robustness score. In this experiment, the value of \( j \) is set to 1.7, where \( j \geq 1 \).

\section{Experiments}

In this section, we demonstrate that our proposed SelfPrompt framework can perform adversarial attacks on large language models such as ChatGPT \cite{OpenAI2024ChatGPT} and Phi-3 \cite{phi32024}, and enable self-evaluation of their robustness based on the results. Additionally, we conduct extensive evaluation experiments on each module within the SelfPrompt framework.

\subsection{Arrangements}

In this subsection, we present the basic arrangements of the experiments, including the datasets used, the large language models employed, and settings of the filter module.

\textbf{Datasets.} We utilize three knowledge graphs (KGs) to generate factual questions: T-REx \cite{elsahar-etal-2018-rex}, which serves as a general-domain KG, and WikiBio \cite{sung2021can} and ULMS \cite{Bodenreider2004TheUM}, which are focused on constrained domains in biology and medicine, respectively. Each predicate in these KGs is paired with a dedicated template, which facilitates template-based original prompt generation within the SelfPrompt framework. For more details about the datasets and their predefined templates, please refer to appendix.

\textbf{Large Language Models.} Our experiments leverage a range of large language models across several series: GPT-4o \cite{OpenAI2024ChatGPT} (including GPT-4o and GPT-4o-mini), Gemma2 \cite{gemma2_2024} (with 2B and 9B parameter versions), Phi-3 \cite{phi32024} (comprising Phi-3-mini with 3.8B parameters and Phi-3-small with 7B parameters), Llama-3.1 \cite{llama2024} (8B parameters), and Mistral \cite{mistral2024} (7B parameters). Variants with different parameter scales within the same model series are employed to examine whether the SelfPrompt framework's evaluation results align with the expectation that "larger models exhibit greater robustness under comparable conditions, particularly when evaluated on general domain datasets", thereby validating the soundness of the evaluation metrics. Meanwhile, models with similar parameter sizes from different series are used to facilitate cross-series comparisons of robustness.

\textbf{Filter Module Setting.} To determine the appropriate values for the two thresholds, \( \tau_t \) and \( \tau_s \), in the filter module, we use a small sample (500 samples per round) generated by various LLMs and different knowledge graph datasets to produce the sentences required for adversarial prompts. We then measure their text fluency and semantic fidelity. The corresponding box plots of the data are presented below.

\begin{figure}[h]
    \centering
    \includegraphics[width=1 \linewidth]{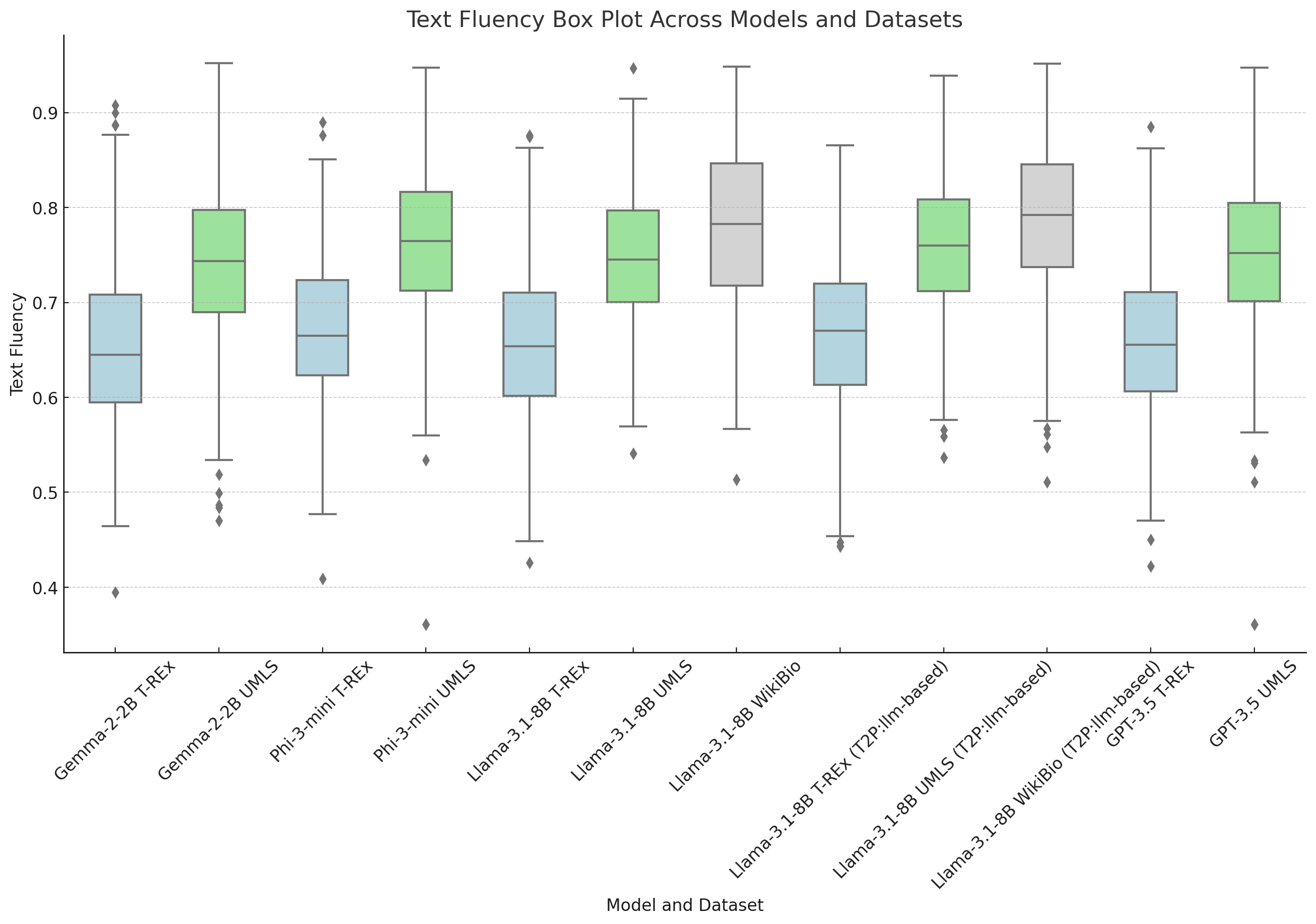}
    \caption{Box Plot of Text Fluency}
    \label{fig:text_fluency}
\end{figure}

\begin{figure}[h]
    \centering
    \includegraphics[width=1 \linewidth]{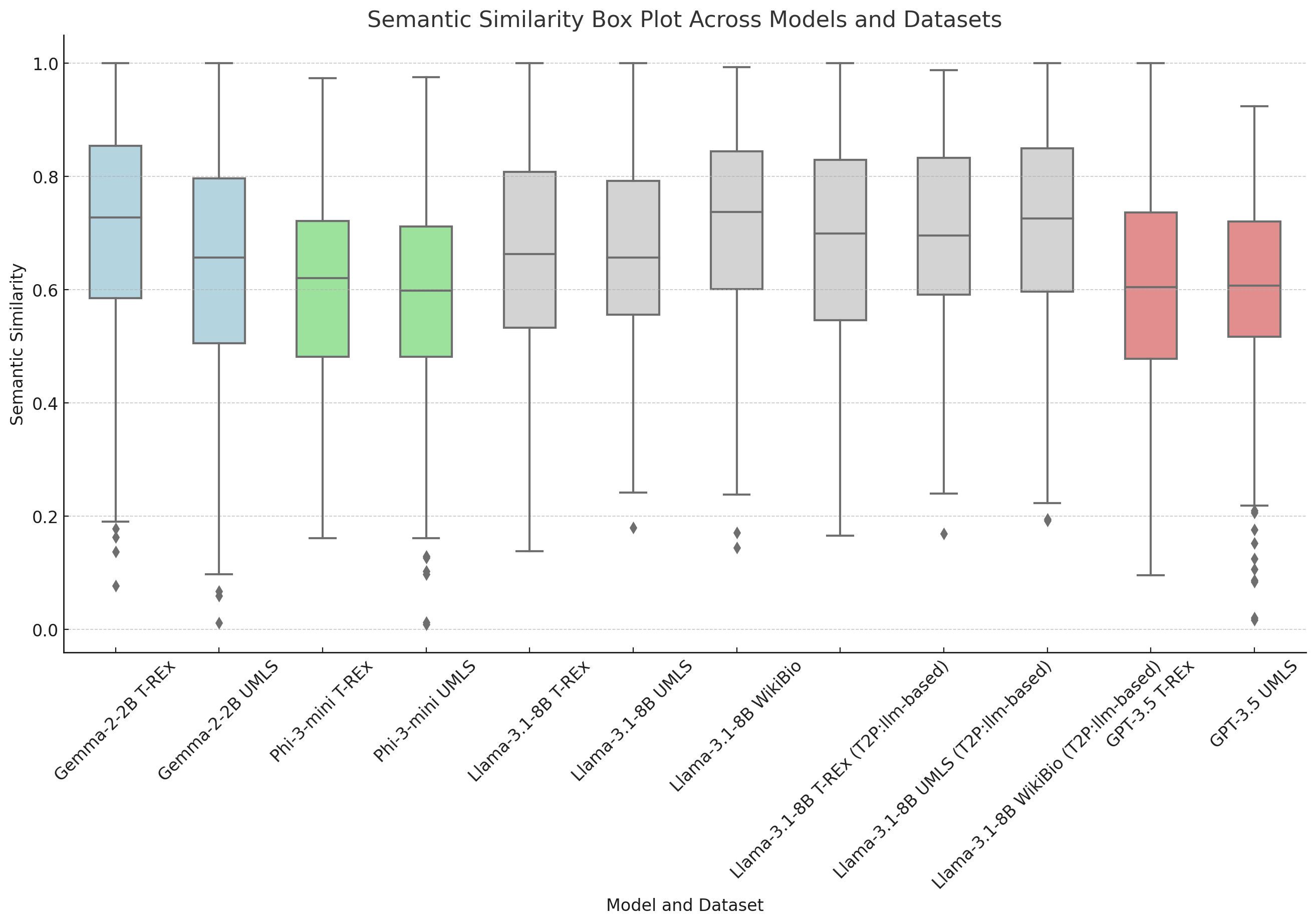}
    \caption{Box Plot of Semantic Fidelity}
    \label{fig:semantic_fidelity}
\end{figure}

In Figure~\ref{fig:text_fluency} and Figure~\ref{fig:semantic_fidelity}, T2P indicates which strategy was used for generating the original prompts. Unless otherwise specified, the template-based strategy is generally applied. As shown in the figures, text fluency is significantly affected by different constrained domains, while semantic fidelity is more influenced by different LLMs; these two metrics are suitable as filtering criteria in the fidelity module. To select high-quality adversarial prompts, we set \( \tau_t = 0.69 \) and \( \tau_s = 0.60 \).

\begin{table*}[t]
\centering
\begin{tabular}{ccccccc}
\toprule
Dataset & Generation Strategy & FS & Gemma2-2B & Gemma2-9B & Phi-3-mini & Phi-3-small \\
\midrule
T-REx & template\_based & No & 0.620 & \textbf{0.631} & 0.607 & \textbf{0.639} \\
T-REx & template\_based & Yes & 0.610 & \textbf{0.641} & 0.639 & \textbf{0.642} \\
T-REx & llm\_based & No & 0.595 & \textbf{0.605} & 0.590 & \textbf{0.647} \\
T-REx & llm\_based & Yes & 0.602 & \textbf{0.633} & 0.584 & \textbf{0.644} \\
UMLS & template\_based & No & \textbf{0.524} & 0.490 & 0.502 & \textbf{0.568} \\
UMLS & template\_based & Yes & 0.500 & \textbf{0.529} & 0.533 & \textbf{0.542} \\
UMLS & llm\_based & No & \textbf{0.510} & 0.459 & \textbf{0.512} & 0.507 \\
UMLS & llm\_based & Yes & \textbf{0.541} & 0.490 & \textbf{0.510} & 0.496 \\
WikiBio & template\_based & No & 0.506 & \textbf{0.555} & 0.502 & \textbf{0.537} \\
WikiBio & template\_based & Yes & \textbf{0.513} & 0.485 & \textbf{0.548} & 0.505 \\
WikiBio & llm\_based & No & \textbf{0.536} & 0.503 & 0.466 & \textbf{0.554} \\
WikiBio & llm\_based & Yes & \textbf{0.508} & 0.501 & \textbf{0.525} & 0.499 \\
\bottomrule
\end{tabular}
\caption{Robustness evaluation results for some models: Gemma2-2B, Gemma2-9B \cite{gemma2_2024}, Phi-3-mini, and Phi-3-small \cite{phi32024}. Bold indicates higher value. The FS column indicates whether the few-shot strategy is used.}
\label{tab:part_robustness_results}
\end{table*}

\subsection{Robustness Evaluation}

Table \ref{tab:part_robustness_results}, Table \ref{tab:part_acc_o}, and Table \ref{tab:part_acc_a} present the robustness evaluation results for selected large models, the accuracy of the LLM on the classification task for set \( \mathcal{O} \) (\( \text{ACC}_{\mathcal{O}} \)), and for set \( \mathcal{A} \) (\( \text{ACC}_{\mathcal{A}} \)), respectively. The test data for the remaining models (including Mistral-7B, Llama-3.1-8B, ChatGPT-4o, and ChatGPT-4o-mini) can be found in the Appendix B.

As shown in Table \ref{tab:part_robustness_results}, when tested on knowledge graph datasets in the general domain, the performance of large language models aligns with the prediction that "within the same series, larger models exhibit greater robustness." This observation validates the effectiveness of our metrics for evaluating model robustness. However, on datasets in constrained domains, the results are not always consistent with this trend. For Phi-3 models, the prediction that larger models are more robust generally holds; even in cases where smaller models show greater robustness, the difference is marginal. In contrast, for the Gemma2 series, smaller models achieve better robustness evaluation results. By comparing Table \ref{tab:part_acc_o} and Table \ref{tab:part_acc_a}, it can be seen that the larger models in the Gemma2 series experience a more significant drop in accuracy when facing adversarial attacks (e.g., for the UMLS dataset, the accuracy drops of the Gemma2-2B and Gemma2-9B models are 0.026 and 0.049, respectively; for the WikiBio dataset, the drops are 0.036 and 0.047, respectively). Thus, the smaller models in the Gemma2 series are less affected by adversarial attacks and therefore demonstrate greater robustness. This could be attributed to the smaller models' limited understanding of specialized domain texts, making them relatively less susceptible to adversarial statements. These findings underscore the necessity of evaluating the robustness of large models in domain-constrained scenarios.

\begin{table*}[t]
\centering
\begin{tabular}{ccccccc}
\toprule
Dataset & Generation Strategy & FS & Gemma2-2B & Gemma2-9B & Phi-3-mini & Phi-3-small \\
\midrule
T-REx & template\_based & No & 0.568 & \textbf{0.622} & 0.560 & \textbf{0.579} \\
T-REx & template\_based & Yes & 0.558 & \textbf{0.609} & 0.527 & \textbf{0.590} \\
T-REx & llm\_based & No & 0.561 & \textbf{0.646} & 0.553 & \textbf{0.612} \\
T-REx & llm\_based & Yes & 0.551 & \textbf{0.654} & 0.514 & \textbf{0.636} \\
UMLS & template\_based & No & 0.429 & \textbf{0.453} & 0.381 & \textbf{0.486} \\
UMLS & template\_based & Yes & \textbf{0.454} & 0.450 & 0.407 & \textbf{0.503} \\
UMLS & llm\_based & No & 0.413 & \textbf{0.416} & \textbf{0.424} & 0.398 \\
UMLS & llm\_based & Yes & \textbf{0.423} & 0.404 & 0.401 & \textbf{0.424} \\
WikiBio & template\_based & No & \textbf{0.462} & 0.430 & \textbf{0.516} & 0.459 \\
WikiBio & template\_based & Yes & \textbf{0.439} & 0.427 & \textbf{0.514} & 0.434 \\
WikiBio & llm\_based & No & 0.422 & \textbf{0.468} & \textbf{0.444} & 0.441 \\
WikiBio & llm\_based & Yes & 0.436 & \textbf{0.472} & \textbf{0.466} & 0.406 \\
\bottomrule
\end{tabular}
\caption{\( \text{ACC}_{\mathcal{O}} \) for some models: Gemma2-2B, Gemma2-9B \cite{gemma2_2024}, Phi-3-mini, and Phi-3-small \cite{phi32024}. Bold indicates higher value. The FS column indicates whether the few-shot strategy is used.}
\label{tab:part_acc_o}
\end{table*}

\begin{table*}[t]
\centering
\begin{tabular}{ccccccc}
\toprule
Dataset & Generation Strategy & FS & Gemma2-2B & Gemma2-9B & Phi-3-mini & Phi-3-small \\
\midrule
T-REx & template\_based & No & 0.549 & \textbf{0.589} & 0.532 & \textbf{0.575} \\
T-REx & template\_based & Yes & 0.534 & \textbf{0.593} & 0.550 & \textbf{0.584} \\
T-REx & llm\_based & No & 0.520 & \textbf{0.574} & 0.512 & \textbf{0.602} \\
T-REx & llm\_based & Yes & 0.523 & \textbf{0.611} & 0.490 & \textbf{0.612} \\
UMLS & template\_based & No & \textbf{0.408} & 0.385 & 0.378 & \textbf{0.465} \\
UMLS & template\_based & Yes & 0.394 & \textbf{0.418} & 0.410 & \textbf{0.446} \\
UMLS & llm\_based & No & \textbf{0.392} & 0.350 & \textbf{0.396} & 0.386 \\
UMLS & llm\_based & Yes & \textbf{0.421} & 0.373 & \textbf{0.389} & 0.383 \\
WikiBio & template\_based & No & 0.401 & \textbf{0.436} & 0.414 & \textbf{0.428} \\
WikiBio & template\_based & Yes & \textbf{0.401} & 0.374 & \textbf{0.456} & 0.393 \\
WikiBio & llm\_based & No & \textbf{0.417} & 0.400 & 0.362 & \textbf{0.438} \\
WikiBio & llm\_based & Yes & 0.396 & \textbf{0.400} & \textbf{0.419} & 0.381 \\
\bottomrule
\end{tabular}
\caption{\( \text{ACC}_{\mathcal{A}} \) for some models: Gemma2-2B, Gemma2-9B \cite{gemma2_2024}, Phi-3-mini, and Phi-3-small \cite{phi32024}. Bold indicates higher value. The FS column indicates whether the few-shot strategy is used.}
\label{tab:part_acc_a}
\end{table*}

\subsection{Experimental Analysis of SelfPrompt}

The robustness evaluation of large language models (LLMs) reveals distinct effects based on the strategies used for generating original prompts (Template-based vs. LLM-based) and whether the few-shot approach is applied in constructing adversarial prompts. Tables \ref{tab:part_robustness_results}, \ref{tab:part_acc_o}, and \ref{tab:part_acc_a} highlight these differences within the same model series.

For generating original prompts, the impact of template-based and LLM-based strategies differs between models in the same series. In the Gemma2 series, the robustness scores for Gemma2-2B and Gemma2-9B under the template-based strategy without few-shot on the T-REx dataset are 0.662 and 0.679, respectively. This suggests that the larger model, Gemma2-9B, benefits slightly from more structured input. However, when using the LLM-based strategy, which introduces more variability, the robustness score for Gemma2-9B on the UMLS dataset drops to 0.530, closer to Gemma2-2B's 0.534. This convergence suggests that more diverse prompts challenge the larger model’s robustness. Table \ref{tab:part_acc_o} shows a similar trend in accuracy \( \text{ACC}_{\mathcal{O}} \), where Gemma2-2B and Gemma2-9B show reduced differences when moving from template-based to LLM-based prompts, highlighting the impact of input variability. In the Phi-3 series, a similar pattern is observed. Under the template-based strategy on the WikiBio dataset, Phi-3-mini and Phi-3-small achieve robustness scores of 0.534 and 0.566, respectively, indicating a benefit for the larger model. However, under the LLM-based strategy, Phi-3-mini's robustness score drops more significantly than Phi-3-small’s (from 0.648 to 0.619 and from 0.695 to 0.694, respectively, on the T-REx dataset), demonstrating that smaller-parameter large language models are relatively weaker than larger-parameter models in generating and understanding natural sentences.

Regarding constructing adversarial prompts, the few-shot Approach significantly affects robustness within model series, as seen in Tables \ref{tab:part_robustness_results} and \ref{tab:part_acc_a}. For the Gemma2 series on the UMLS dataset, Gemma2-9B's robustness drops from 0.529 without few-shot to 0.490 with few-shot, revealing increased vulnerability under adversarial conditions. In contrast, Gemma2-2B shows a smaller drop (from 0.500 to 0.512), indicating less sensitivity to adversarial prompts. In the Phi-3 series, on the WikiBio dataset, Phi-3-mini's accuracy \( \text{ACC}_{\mathcal{A}} \) drops significantly from 0.612 to 0.521 when few-shot is applied, compared to a smaller decrease for Phi-3-small. This highlights the effectiveness of few-shot in generating more challenging adversarial prompts that test model robustness.

In summary, the choice of strategy for generating original prompts and constructing adversarial prompts significantly influences the robustness evaluation of LLMs. Template-based strategies offer a controlled environment that favors larger models, while LLM-based strategies and the few-shot approach introduce more variability and difficulty, providing a more comprehensive robustness assessment within the same model series.

\section{Conclusion}

This paper introduces SelfPrompt, a framework for autonomously evaluating the robustness of large language models (LLMs) using domain-constrained knowledge guidelines and refined adversarial prompts. Our experiments confirm that the proposed method provides a reliable and effective evaluation of LLM robustness across various domains, demonstrating that larger models generally show greater robustness in general settings, while results may vary in domain-specific scenarios. Future work could explore expanding this framework to cover more diverse knowledge graphs and adaptive prompt generation techniques.

\section*{Limitations}

The limitations of our work includes:

\begin{itemize}

\item Types of problems for evaluating LLM robustness. In the SelfPrompt framework, we require the LLM to perform classification tasks to evaluate the robustness of large language models; in future research, we plan to enrich the types of problems by including types such as short answer questions and true/false questions, to conduct a more comprehensive evaluation of the LLM of robustness.

\item The SelfPrompt framework relies on existing knowledge graphs. When suitable knowledge graphs are lacking in a specific domain, constructing such knowledge graphs for that domain increases the usage cost of this framework. In future experiments, we plan to attempt constructing a small number of triplets directly without relying on knowledge graphs, for robustness evaluation purposes.

\item Lack of further comparative experiments. It is due to the unique design of the robustness evaluation metrics introduced in this paper, which limits the ability to compare with existing robustness evaluation frameworks. In future experiments, we plan to conduct further comparative tests once similar frameworks become available.

\end{itemize}

\bibliography{custom}

\begin{thebibliography}{41}
\providecommand{\natexlab}[1]{#1}

\bibitem[{Ailem et~al.(2024)Ailem, Marazopoulou, Siska, and Bono}]{ailem2024examining}
Melissa Ailem, Katerina Marazopoulou, Charlotte Siska, and James Bono. 2024.
\newblock Examining the robustness of llm evaluation to the distributional assumptions of benchmarks.
\newblock \emph{arXiv preprint arXiv:2404.16966}.

\bibitem[{Axelsson and Skantze(2023)}]{axelsson2023using}
Agnes Axelsson and Gabriel Skantze. 2023.
\newblock Using large language models for zero-shot natural language generation from knowledge graphs.
\newblock \emph{arXiv preprint arXiv:2307.07312}.

\bibitem[{Bodenreider(2004)}]{Bodenreider2004TheUM}
Olivier Bodenreider. 2004.
\newblock The unified medical language system (umls): integrating biomedical terminology.
\newblock \emph{Nucleic acids research}, 32(suppl\_1):D267--D270.

\bibitem[{Brown et~al.(2020)Brown, Mann, Ryder, Subbiah, Kaplan, Dhariwal, Neelakantan, Shyam, Sastry, Askell et~al.}]{brown2020language}
Tom Brown, Benjamin Mann, Nick Ryder, Melanie Subbiah, Jared~D Kaplan, Prafulla Dhariwal, Arvind Neelakantan, Pranav Shyam, Girish Sastry, Amanda Askell, et~al. 2020.
\newblock Language models are few-shot learners.
\newblock \emph{Advances in neural information processing systems}, 33:1877--1901.

\bibitem[{Chen et~al.(2023)Chen, Wu, and Zaki}]{chen2023toward}
Yu~Chen, Lingfei Wu, and Mohammed~J Zaki. 2023.
\newblock Toward subgraph-guided knowledge graph question generation with graph neural networks.
\newblock \emph{IEEE Transactions on Neural Networks and Learning Systems}.

\bibitem[{Chiang et~al.(2023)Chiang, Li, Lin, Sheng, Wu, Zhang, Zheng, Zhuang, Zhuang, Gonzalez, Stoica, and Xing}]{vicuna2023}
Wei-Lin Chiang, Zhuohan Li, Zi~Lin, Ying Sheng, Zhanghao Wu, Hao Zhang, Lianmin Zheng, Siyuan Zhuang, Yonghao Zhuang, Joseph~E. Gonzalez, Ion Stoica, and Eric~P. Xing. 2023.
\newblock \href {https://lmsys.org/blog/2023-03-30-vicuna/} {Vicuna: An open-source chatbot impressing gpt-4 with 90\%* chatgpt quality}.

\bibitem[{Elsahar et~al.(2018)Elsahar, Vougiouklis, Remaci, Gravier, Hare, Laforest, and Simperl}]{elsahar-etal-2018-rex}
Hady Elsahar, Pavlos Vougiouklis, Arslen Remaci, Christophe Gravier, Jonathon Hare, Frederique Laforest, and Elena Simperl. 2018.
\newblock T-rex: A large scale alignment of natural language with knowledge base triples.
\newblock In \emph{Proceedings of the Eleventh International Conference on Language Resources and Evaluation (LREC 2018)}.

\bibitem[{Gemma~Team(2024)}]{gemma2_2024}
Google~DeepMind Gemma~Team. 2024.
\newblock \href {https://arxiv.org/abs/2408.00118} {Gemma 2: Improving open language models at a practical size}.
\newblock \emph{arXiv preprint arXiv:2408.00118}.

\bibitem[{Goel et~al.(2021)Goel, Rajani, Vig, Tan, Wu, Zheng, Xiong, Bansal, and R{\'e}}]{goel2021robustness}
Karan Goel, Nazneen Rajani, Jesse Vig, Samson Tan, Jason Wu, Stephan Zheng, Caiming Xiong, Mohit Bansal, and Christopher R{\'e}. 2021.
\newblock Robustness gym: Unifying the nlp evaluation landscape.
\newblock \emph{arXiv preprint arXiv:2101.04840}.

\bibitem[{Goodfellow et~al.(2016)Goodfellow, Bengio, and Courville}]{Goodfellow-et-al-2016}
Ian Goodfellow, Yoshua Bengio, and Aaron Courville. 2016.
\newblock \href {http://www.deeplearningbook.org} {\emph{Deep Learning}}.
\newblock MIT Press.

\bibitem[{Guo et~al.(2022)Guo, Zhang, Wang, Zhang, Li, and Chen}]{guo2022dsm}
Shasha Guo, Jing Zhang, Yanling Wang, Qianyi Zhang, Cuiping Li, and Hong Chen. 2022.
\newblock Dsm: Question generation over knowledge base via modeling diverse subgraphs with meta-learner.
\newblock In \emph{Proceedings of the 2022 Conference on Empirical Methods in Natural Language Processing}, pages 4194--4207.

\bibitem[{Gururangan et~al.(2018)Gururangan, Swayamdipta, Levy, Schwartz, Bowman, and Smith}]{gururangan2018annotation}
Suchin Gururangan, Swabha Swayamdipta, Omer Levy, Roy Schwartz, Samuel Bowman, and Noah~A Smith. 2018.
\newblock Annotation artifacts in natural language inference data.
\newblock In \emph{Proceedings of the 2018 Conference of the North American Chapter of the Association for Computational Linguistics: Human Language Technologies, Volume 2 (Short Papers)}, pages 107--112.

\bibitem[{Kumar et~al.(2019)Kumar, Hua, Ramakrishnan, Qi, Gao, and Li}]{kumar2019difficulty}
Vishwajeet Kumar, Yuncheng Hua, Ganesh Ramakrishnan, Guilin Qi, Lianli Gao, and Yuan-Fang Li. 2019.
\newblock Difficulty-controllable multi-hop question generation from knowledge graphs.
\newblock In \emph{The Semantic Web--ISWC 2019: 18th International Semantic Web Conference, Auckland, New Zealand, October 26--30, 2019, Proceedings, Part I 18}, pages 382--398. Springer.

\bibitem[{Le~Bras et~al.(2020)Le~Bras, Swayamdipta, Bhagavatula, Zellers, Peters, Sabharwal, and Choi}]{le2020adversarial}
Ronan Le~Bras, Swabha Swayamdipta, Chandra Bhagavatula, Rowan Zellers, Matthew Peters, Ashish Sabharwal, and Yejin Choi. 2020.
\newblock Adversarial filters of dataset biases.
\newblock In \emph{International conference on machine learning}, pages 1078--1088. Pmlr.

\bibitem[{Li et~al.(2023)Li, Liu, Gao, and Buntine}]{li2023survey}
Xinzhe Li, Ming Liu, Shang Gao, and Wray Buntine. 2023.
\newblock A survey on out-of-distribution evaluation of neural nlp models.
\newblock \emph{arXiv preprint arXiv:2306.15261}.

\bibitem[{Logan~IV et~al.(2021)Logan~IV, Bala{\v{z}}evi{\'c}, Wallace, Petroni, Singh, and Riedel}]{logan2021cutting}
Robert~L Logan~IV, Ivana Bala{\v{z}}evi{\'c}, Eric Wallace, Fabio Petroni, Sameer Singh, and Sebastian Riedel. 2021.
\newblock Cutting down on prompts and parameters: Simple few-shot learning with language models.
\newblock \emph{arXiv preprint arXiv:2106.13353}.

\bibitem[{Luo et~al.(2024)Luo, Ghawanmeh, Zhu, and Khattak}]{luo2024biaskg}
Chu~Fei Luo, Ahmad Ghawanmeh, Xiaodan Zhu, and Faiza~Khan Khattak. 2024.
\newblock Biaskg: Adversarial knowledge graphs to induce bias in large language models.
\newblock \emph{arXiv preprint arXiv:2405.04756}.

\bibitem[{Luo et~al.(2023)Luo, Vu, Phung, and Haffari}]{luo2023systematic}
Linhao Luo, Thuy-Trang Vu, Dinh Phung, and Gholamreza Haffari. 2023.
\newblock Systematic assessment of factual knowledge in large language models.
\newblock \emph{arXiv preprint arXiv:2310.11638}.

\bibitem[{Manning et~al.(2008)Manning, Raghavan, and Sch{\"u}tze}]{Manning:2008:IIR:1394399}
Christopher~D. Manning, Prabhakar Raghavan, and Hinrich Sch{\"u}tze. 2008.
\newblock \emph{Introduction to Information Retrieval}.
\newblock Cambridge University Press, New York, NY, USA.

\bibitem[{Meng et~al.(2024)Meng, Shao, Wang, Qiao, and Xu}]{meng2024advancing}
Lingzhuang Meng, Mingwen Shao, Fan Wang, Yuanjian Qiao, and Zhaofei Xu. 2024.
\newblock Advancing few-shot black-box attack with alternating training.
\newblock \emph{IEEE Transactions on Reliability}.

\bibitem[{Mistral and contributors(2024)}]{mistral2024}
Team Mistral and contributors. 2024.
\newblock \href {https://arxiv.org/abs/2406.09876} {Mistral: A multi-purpose language model for comprehensive text understanding}.
\newblock \emph{arXiv preprint arXiv:2406.09876}.
\newblock Accessed: 2024-09-14.

\bibitem[{Mizrahi et~al.(2023)Mizrahi, Kaplan, Malkin, Dror, Shahaf, and Stanovsky}]{mizrahi2023state}
Moran Mizrahi, Guy Kaplan, Dan Malkin, Rotem Dror, Dafna Shahaf, and Gabriel Stanovsky. 2023.
\newblock State of what art? a call for multi-prompt llm evaluation.
\newblock \emph{arXiv preprint arXiv:2401.00595}.

\bibitem[{Nie et~al.(2020)Nie, Williams, Dinan, Bansal, Weston, and Kiela}]{nie2020adversarial}
Yixin Nie, Adina Williams, Emily Dinan, Mohit Bansal, Jason Weston, and Douwe Kiela. 2020.
\newblock Adversarial nli: A new benchmark for natural language understanding.
\newblock In \emph{Proceedings of the 58th Annual Meeting of the Association for Computational Linguistics}, pages 4885--4901.

\bibitem[{Niven and Kao(2019)}]{niven2019probing}
Timothy Niven and Hung-Yu Kao. 2019.
\newblock Probing neural network comprehension of natural language arguments.
\newblock In \emph{Proceedings of the 57th Annual Meeting of the Association for Computational Linguistics}, pages 4658--4664.

\bibitem[{OpenAI(2024)}]{OpenAI2024ChatGPT}
OpenAI. 2024.
\newblock \href {https://www.openai.com/} {Chatgpt: A large language model by openai}.
\newblock Accessed: 2024-09-14.

\bibitem[{Research(2024)}]{phi32024}
Microsoft Research. 2024.
\newblock \href {https://www.microsoft.com/research/project/phi-3/} {Phi-3: Advanced instruction-following language model}.
\newblock Accessed: 2024-09-14.

\bibitem[{Seyler et~al.(2017)Seyler, Yahya, and Berberich}]{seyler2017knowledge}
Dominic Seyler, Mohamed Yahya, and Klaus Berberich. 2017.
\newblock Knowledge questions from knowledge graphs.
\newblock In \emph{Proceedings of the ACM SIGIR international conference on theory of information retrieval}, pages 11--18.

\bibitem[{Sung et~al.(2021)Sung, Lee, Yi, Jeon, Kim, and Kang}]{sung2021can}
Mujeen Sung, Jinhyuk Lee, Sean Yi, Minji Jeon, Sungdong Kim, and Jaewoo Kang. 2021.
\newblock Can language models be biomedical knowledge bases?
\newblock \emph{arXiv preprint arXiv:2109.07154}.

\bibitem[{Taori et~al.(2023)Taori, Gulrajani, Zhang, Dubois, Li, Guestrin, Liang, and Hashimoto}]{taori2023alpaca}
Rohan Taori, Ishaan Gulrajani, Tianyi Zhang, Yann Dubois, Xuechen Li, Carlos Guestrin, Percy Liang, and Tatsunori~B Hashimoto. 2023.
\newblock Alpaca: A strong, replicable instruction-following model.
\newblock \emph{Stanford Center for Research on Foundation Models. https://crfm. stanford. edu/2023/03/13/alpaca. html}, 3(6):7.

\bibitem[{Touvron et~al.(2023)Touvron, Lavril, Izacard, Martinet, Lachaux, Lacroix, Rozi{\`e}re, Goyal, Hambro, Azhar et~al.}]{touvron2023llama}
Hugo Touvron, Thibaut Lavril, Gautier Izacard, Xavier Martinet, Marie-Anne Lachaux, Timoth{\'e}e Lacroix, Baptiste Rozi{\`e}re, Naman Goyal, Eric Hambro, Faisal Azhar, et~al. 2023.
\newblock Llama: Open and efficient foundation language models.
\newblock \emph{arXiv preprint arXiv:2302.13971}.

\bibitem[{Touvron et~al.(2024)Touvron, Lavril, Izacard et~al.}]{llama2024}
Hugo Touvron, Thibaut Lavril, Gautier Izacard, et~al. 2024.
\newblock \href {https://arxiv.org/abs/2404.12345} {Llama-3.1: Efficient and scalable foundation language models}.
\newblock \emph{arXiv preprint arXiv:2404.12345}.
\newblock Accessed: 2024-09-14.

\bibitem[{Voronov et~al.(2024)Voronov, Wolf, and Ryabinin}]{voronov2024mind}
Anton Voronov, Lena Wolf, and Max Ryabinin. 2024.
\newblock Mind your format: Towards consistent evaluation of in-context learning improvements.
\newblock \emph{arXiv preprint arXiv:2401.06766}.

\bibitem[{Wang et~al.(2018)Wang, Singh, Michael, Hill, Levy, and Bowman}]{wang2018glue}
Alex Wang, Amanpreet Singh, Julian Michael, Felix Hill, Omer Levy, and Samuel~R Bowman. 2018.
\newblock Glue: A multi-task benchmark and analysis platform for natural language understanding.
\newblock \emph{arXiv preprint arXiv:1804.07461}.

\bibitem[{Wang et~al.(2023{\natexlab{a}})Wang, Chen, Pei, Xie, Kang, Zhang, Xu, Xiong, Dutta, Schaeffer et~al.}]{wang2023decodingtrust}
Boxin Wang, Weixin Chen, Hengzhi Pei, Chulin Xie, Mintong Kang, Chenhui Zhang, Chejian Xu, Zidi Xiong, Ritik Dutta, Rylan Schaeffer, et~al. 2023{\natexlab{a}}.
\newblock Decodingtrust: A comprehensive assessment of trustworthiness in gpt models.
\newblock \emph{arXiv preprint arXiv:2306.11698}.

\bibitem[{Wang et~al.(2021)Wang, Xu, Wang, Gan, Cheng, Gao, Awadallah, and Li}]{wang2021adversarial}
Boxin Wang, Chejian Xu, Shuohang Wang, Zhe Gan, Yu~Cheng, Jianfeng Gao, Ahmed~Hassan Awadallah, and Bo~Li. 2021.
\newblock Adversarial glue: A multi-task benchmark for robustness evaluation of language models.
\newblock \emph{arXiv preprint arXiv:2111.02840}.

\bibitem[{Wang et~al.(2023{\natexlab{b}})Wang, Hu, Hou, Chen, Zheng, Wang, Yang, Huang, Ye, Geng et~al.}]{wang2023robustness}
Jindong Wang, Xixu Hu, Wenxin Hou, Hao Chen, Runkai Zheng, Yidong Wang, Linyi Yang, Haojun Huang, Wei Ye, Xiubo Geng, et~al. 2023{\natexlab{b}}.
\newblock On the robustness of chatgpt: An adversarial and out-of-distribution perspective.
\newblock \emph{arXiv preprint arXiv:2302.12095}.

\bibitem[{Weber et~al.(2023)Weber, Bruni, and Hupkes}]{weber2023mind}
Lucas Weber, Elia Bruni, and Dieuwke Hupkes. 2023.
\newblock Mind the instructions: a holistic evaluation of consistency and interactions in prompt-based learning.
\newblock \emph{arXiv preprint arXiv:2310.13486}.

\bibitem[{Xu et~al.(2023)Xu, Kong, Liu, Cui, Wang, Zhang, and Kankanhalli}]{xu2023llm}
Xilie Xu, Keyi Kong, Ning Liu, Lizhen Cui, Di~Wang, Jingfeng Zhang, and Mohan Kankanhalli. 2023.
\newblock An llm can fool itself: A prompt-based adversarial attack.
\newblock \emph{arXiv preprint arXiv:2310.13345}.

\bibitem[{Yang et~al.(2023)Yang, Zhang, Qin, Li, Wang, Liu, Wang, Xie, and Zhang}]{yang2023glue}
Linyi Yang, Shuibai Zhang, Libo Qin, Yafu Li, Yidong Wang, Hanmeng Liu, Jindong Wang, Xing Xie, and Yue Zhang. 2023.
\newblock Glue-x: Evaluating natural language understanding models from an out-of-distribution generalization perspective.
\newblock In \emph{Findings of the Association for Computational Linguistics: ACL 2023}, pages 12731--12750.

\bibitem[{Zhu et~al.(2023)Zhu, Wang, Zhou, Wang, Chen, Wang, Yang, Ye, Zhang, Gong et~al.}]{zhu2023promptbench}
Kaijie Zhu, Jindong Wang, Jiaheng Zhou, Zichen Wang, Hao Chen, Yidong Wang, Linyi Yang, Wei Ye, Yue Zhang, Neil~Zhenqiang Gong, et~al. 2023.
\newblock Promptbench: Towards evaluating the robustness of large language models on adversarial prompts.
\newblock \emph{arXiv preprint arXiv:2306.04528}.

\bibitem[{Zhuo et~al.(2023)Zhuo, Li, Huang, Shiri, Wang, Haffari, and Li}]{zhuo2023robustness}
Terry~Yue Zhuo, Zhuang Li, Yujin Huang, Fatemeh Shiri, Weiqing Wang, Gholamreza Haffari, and Yuan-Fang Li. 2023.
\newblock On robustness of prompt-based semantic parsing with large pre-trained language model: An empirical study on codex.
\newblock In \emph{Proceedings of the 17th Conference of the European Chapter of the Association for Computational Linguistics}, pages 1090--1102.

\end{thebibliography}

\appendix

\clearpage 
\section{Experimentation Details}

\subsection{Dataset}

In this experiment, we divide the knowledge graph dataset into two categories based on the domain of knowledge represented by the knowledge graphs, including the general domain knowledge graphs and the constrained domain knowledge graphs. The general domain knowledge graph datasets is T-REx; the constrained domain knowledge graph datasets include UMLS and WikiBio.

\begin{itemize}

\item \textbf{T-REx.} \cite{elsahar-etal-2018-rex} Originating from Wikipedia, this is a general domain knowledge graph that records a large number of triplets belonging to various fields.
\item \textbf{UMLS.} \cite{Bodenreider2004TheUM} This is a constrained-domain knowledge graph in the medical field, constructed by experts in the domain, and it contains information about various medical concepts and their relationships.
\item \textbf{WikiBio.} \cite{sung2021can} This dataset is constructed by extracting biological instances from Wikidata and is a constrained-domain knowledge graph in the field of biology.

\end{itemize}

\subsection{Loss Function and Cosine Similarity Used in Filter Module}

\textbf{Loss Function.} The loss function is a mathematical function that measures the difference between the predicted outputs of a model and the actual outputs (ground truth). The Cross-Entropy Loss is commonly used in the context of language models. It is defined as:

\begin{equation}
\text{Loss}(s) = - \sum_{i=1}^{N} \log P(x_i \,|\, x_{<i})
\end{equation}

where $x_i$ is the $i$-th token in a sequence, and $P(x_i \,|\, x_{<i})$ is the conditional probability of the token given all previous tokens \cite{Goodfellow-et-al-2016}.

\textbf{Cosine Similarity.} Cosine similarity is a metric used to measure how similar two vectors are, irrespective of their magnitude. It is often used in natural language processing for comparing the similarity between text embeddings. The cosine similarity between vectors $A$ and $B$ is defined as:

\begin{equation}
\text{Cosine Similarity}(A, B) = \frac{A \cdot B}{\|A\| \|B\|}
\end{equation}

where $A \cdot B$ is the dot product of vectors $A$ and $B$, and $\|A\|$ and $\|B\|$ are the magnitudes (norms) of vectors $A$ and $B$. This metric ranges from $-1$ to $1$, where $1$ indicates that the vectors are identical, $0$ means they are orthogonal (dissimilar), and $-1$ means they are diametrically opposed. \cite{Manning:2008:IIR:1394399}.

\subsection{Implementations}

\textbf{Large Language Model.} We utilize several models from the ChatGPT family \cite{OpenAI2024ChatGPT}, including GPT-4o and GPT-4o-mini. The large language models were accessed via paid APIs to complete relevant robustness evaluation tasks. We also used several open-source models, including Gemma2 \cite{gemma2_2024} (with 2B and 9B parameter versions), Phi-3 \cite{phi32024} (comprising Phi-3-mini with 3.8B parameters and Phi-3-small with 7B parameters), Llama-3.1 \cite{llama2024} (8B parameters), and Mistral \cite{mistral2024} (7B parameters). These open-source models were run locally with FP16 precision on a single RTX-4090 GPU.

\noindent{}\textbf{Prompt Generation and Response Processing.} We set the ratio of the three labels "true," "entity\_error," and "predicate\_error" for the generated prompts to 1:1:1. To extract the classification results from responses of the LLM for the classification task, we employed string matching. If a response matches one of the aforementioned three labels and the label is the correct one, classification of the LLM is deemed correct; otherwise, it is considered incorrect. For each large model on each knowledge graph dataset, we generated 1,000 adversarial prompts for experiments under each specific condition of the original prompt generation strategy and the few-shot strategy.

\section{Partial Experimental Results}

This subsection presents partial experimental results. It includes the values of \( \text{ACC}_{\mathcal{O}} \) and \( \text{ACC}_{\mathcal{A}} \), as well as robustness evaluation results for adversarial attacks on Llama-3.1, Mistral, ChatGPT-4o, and ChatGPT-4o-mini. The detailed results are presented in Tables \ref{tab:part_robustness_results_1}, \ref{tab:part_acc_o_1}, and \ref{tab:part_acc_a_1}. As shown in the tables, Llama-3.1 exhibits poor robustness, significantly lagging behind the Mistral model of the same parameter size. Additionally, GPT-4o-mini demonstrates better robustness than GPT-4o, which could be attributed to its later release and the subsequent improvements in robustness.

\begin{table*}[h]
\centering
\begin{tabular}{ccccccc}
\toprule
Dataset & Generation Strategy & FS & Llama-3.1 & Mistral & ChatGPT-4o-mini & ChatGPT-4o \\
\midrule
T-REx & template\_based & No & 0.404 & 0.589 & 0.661 & 0.496 \\
T-REx & template\_based & Yes & 0.418 & 0.585 & 0.660 & 0.568 \\
T-REx & llm\_based & No & 0.417 & 0.496 & 0.633 & 0.508 \\
T-REx & llm\_based & Yes & 0.474 & 0.507 & 0.646 & 0.516 \\
UMLS & template\_based & No & 0.437 & 0.523 & 0.565 & 0.535 \\
UMLS & template\_based & Yes & 0.465 & 0.564 & 0.530 & 0.492 \\
UMLS & llm\_based & No & 0.510 & 0.466 & 0.542 & 0.509 \\
UMLS & llm\_based & Yes & 0.541 & 0.490 & 0.566 & 0.496 \\
WikiBio & template\_based & No & 0.475 & 0.532 & 0.587 & 0.494 \\
WikiBio & template\_based & Yes & 0.483 & 0.548 & 0.573 & 0.512 \\
WikiBio & llm\_based & No & 0.513 & 0.486 & 0.621 & 0.501 \\
WikiBio & llm\_based & Yes & 0.495 & 0.487 & 0.637 & 0.509 \\
\bottomrule
\end{tabular}
\caption{Robustness Evaluation Results for Some Models: Llama-3.1 \cite{llama2024}, Mistral \cite{mistral2024}, ChatGPT-4o-mini, and ChatGPT-4o \cite{OpenAI2024ChatGPT}. The FS column indicates whether the Few-shot Strategy is used.}
\label{tab:part_robustness_results_1}
\end{table*}

\begin{table*}[h]
\centering
\begin{tabular}{ccccccc}
\toprule
Dataset & Generation Strategy & FS & Llama-3.1 & Mistral & ChatGPT-4o-mini & ChatGPT-4o \\
\midrule
T-REx & template\_based & No & 0.302 & 0.503 & 0.603 & 0.583 \\
T-REx & template\_based & Yes & 0.306 & 0.521 & 0.621 & 0.631 \\
T-REx & llm\_based & No & 0.316 & 0.528 & 0.591 & 0.606 \\
T-REx & llm\_based & Yes & 0.338 & 0.562 & 0.596 & 0.648 \\
UMLS & template\_based & No & 0.321 & 0.442 & 0.502 & 0.491 \\
UMLS & template\_based & Yes & 0.325 & 0.470 & 0.514 & 0.489 \\
UMLS & llm\_based & No & 0.333 & 0.453 & 0.474 & 0.462 \\
UMLS & llm\_based & Yes & 0.344 & 0.466 & 0.483 & 0.501 \\
WikiBio & template\_based & No & 0.315 & 0.433 & 0.475 & 0.444 \\
WikiBio & template\_based & Yes & 0.307 & 0.441 & 0.493 & 0.462 \\
WikiBio & llm\_based & No & 0.299 & 0.472 & 0.508 & 0.491 \\
WikiBio & llm\_based & Yes & 0.289 & 0.483 & 0.495 & 0.478 \\
\bottomrule
\end{tabular}
\caption{\( \text{ACC}_{\mathcal{O}} \) for Some Models: Llama-3.1 \cite{llama2024}, Mistral \cite{mistral2024}, ChatGPT-4o-mini, and ChatGPT-4o \cite{OpenAI2024ChatGPT}. The FS column indicates whether the Few-shot Strategy is used.}
\label{tab:part_acc_o_1}
\end{table*}

\begin{table*}[h]
\centering
\begin{tabular}{ccccccc}
\toprule
Dataset & Generation Strategy & FS & Llama-3.1 & Mistral & ChatGPT-4o-mini & ChatGPT-4o \\
\midrule
T-REx & template\_based & No & 0.287 & 0.491 & 0.612 & 0.432 \\
T-REx & template\_based & Yes & 0.298 & 0.490 & 0.621 & 0.526 \\
T-REx & llm\_based & No & 0.299 & 0.412 & 0.575 & 0.453 \\
T-REx & llm\_based & Yes & 0.347 & 0.434 & 0.592 & 0.480 \\
UMLS & template\_based & No & 0.315 & 0.411 & 0.467 & 0.436 \\
UMLS & template\_based & Yes & 0.346 & 0.416 & 0.476 & 0.431 \\
UMLS & llm\_based & No & 0.324 & 0.424 & 0.482 & 0.441 \\
UMLS & llm\_based & Yes & 0.321 & 0.438 & 0.495 & 0.459 \\
WikiBio & template\_based & No & 0.299 & 0.414 & 0.523 & 0.482 \\
WikiBio & template\_based & Yes & 0.310 & 0.428 & 0.506 & 0.499 \\
WikiBio & llm\_based & No & 0.312 & 0.439 & 0.536 & 0.501 \\
WikiBio & llm\_based & Yes & 0.321 & 0.417 & 0.518 & 0.486 \\
\bottomrule
\end{tabular}
\caption{\( \text{ACC}_{\mathcal{A}} \) for Some Models: Llama-3.1 \cite{llama2024}, Mistral \cite{mistral2024}, ChatGPT-4o-mini, and ChatGPT-4o \cite{OpenAI2024ChatGPT}. The FS column indicates whether the Few-shot Strategy is used.}
\label{tab:part_acc_a_1}
\end{table*}

\section{Prompt Templates}

In this section, we introduce the prompt templates used in the SelfPrompt framework. These prompt templates include: the Triplets-to-Prompts Template for generating original prompts when selecting the LLM-based strategy; the Adversarial Prompts Generation Template for constructing adversarial prompts; the Examples-Generation Template for generating prompt examples required when using the few-shot strategy; and the (Non-)Adversarial Prompt Template for generating prompts and requiring LLMs to classify the label of the sentence in the prompts.

\subsection{Triplets-to-Prompts Template}

This template is responsible for transforming a triplet formatted as $t = (s, p, o) \in \mathcal{D}$, where $s$ denotes the subject of the triplet, and $p$ and $o$ refer to the predicate and object of the triplet, respectively. The template converts this triplet into a naturally described sentence, where the positions marked in red in the template need to be replaced with the content of the triplets.

\begin{tcolorbox}[colback=gray!5!white, colframe=black, title=Triplets-to-Prompts Template, breakable]
Here is a triple (subject, predicate, object) extracted from a knowledge graph:
\begin{itemize}
    \item Subject(s): \{\textcolor{red}{subject}\}
    \item Subject Alias(es): \{\textcolor{red}{subject alias}\}
    \item Predicate: \{\textcolor{red}{predicate}\}
    \item Template of the Predicate: \{\textcolor{red}{predicate template}\}
    \item Description of the Predicate: \{\textcolor{red}{predicate description}\}
    \item Object(s): \{\textcolor{red}{object}\}
    \item Object Alias(es): \{\textcolor{red}{object alias}\}
\end{itemize}

Please create a statement describing this triple.

\textbf{Note:}
\begin{itemize}
    \item The truthfulness of the triple is not important.
    \item Do not alter the meaning of the predicate.
\end{itemize}

\textbf{Statement:}
\end{tcolorbox}

\subsection{Adversarial Prompts Generation Template}

This template is designed to transform original prompts into adversarial prompts. It requires the provision of a sentence from the original prompt that describes the corresponding triplet, along with its constituent components. When employing a few-shot strategy, this template also necessitates the inclusion of corresponding examples.

\begin{tcolorbox}[colback=gray!5!white, colframe=black, title=Adversarial Prompts Generation Template, breakable]
You are given a knowledge graph triplet and a sentence generated from this triplet.
Your task is to paraphrase the sentence while keeping the semantic meaning of the new sentence unchanged.
The paraphrased sentence should be classified into a different label from the current one.
Use the given information about subjects, objects, their aliases, and the predicate to guide your paraphrasing.

Here are the detailed steps for the task:

1. \textbf{Paraphrase the Sentence}:
   \begin{itemize}
       \item Rewrite the given sentence in a different way.
       \item Ensure that the rewritten sentence maintains the same semantic meaning as the original sentence.
   \end{itemize}

2. \textbf{Change the Classification}:
   \begin{itemize}
       \item The new paraphrased sentence should be classified into a different label from the current label.
       \item The possible labels are ["true", "entity\_error", "predicate\_error"].
       \item true: The triplet and the sentence accurately reflect the true content.
       \item entity\_error: The triplet contains an error related to the Subject or Object, as well as the sentence.
       \item predicate\_error: The triplet contains an error related to the Predicate, as well as in the sentence.
   \end{itemize}

(Here is five examples that fit the guidance:
Original Sentence: \{\textcolor{red}{original sentence 1}\} -> Paraphrased Sentence: \{\textcolor{red}{paraphrased sentence 1}\} \\
... \\
Original Sentence: \{\textcolor{red}{original sentence 5}\} -> Paraphrased Sentence: \{\textcolor{red}{paraphrased sentence 5}\})

Here is the input triplet and sentence for you to paraphrase:

\textbf{Triplet:}
\{
\{ "subs": \{\textcolor{red}{subject}\}, "sub\_aliases": \{\textcolor{red}{subject alias}\}, "predicate": \{ "predicate": \{\textcolor{red}{predicate}\}, "template": \{\textcolor{red}{predicate template}\}, "description": \{\textcolor{red}{predicate description}\} \}, "objs": \{\textcolor{red}{object}\}, "obj\_aliases": \{\textcolor{red}{object alias}\}, "label": \{\textcolor{red}{label}\} \}
\}

Where:
\begin{itemize}
    \item "subs" is a list of subjects.
    \item "sub\_aliases" is a list of subject aliases.
    \item "predicate" is an object containing the name, template, and description of the relationship.
    \item "objs" is a list of objects.
    \item "obj\_aliases" is a list of object aliases.
    \item "label" is the current label of the triplet, which is one of ["true", "entity\_error", "predicate\_error"].
\end{itemize}

\textbf{Sentence:} \{\textcolor{red}{sentence}\}

\textbf{Important:} The classification of this sentence is: '\{\textcolor{red}{sentence label}\}', and label of the sentence after paraphrase should be '\{\textcolor{red}{wrong label 1}\}' or '\{\textcolor{red}{wrong label 2}\}'.

Only output the paraphrased sentence without anything else. \\
\textbf{Paraphrased Sentence:}
\end{tcolorbox}

\subsection{Examples-Generation Template}

This template is responsible for generating example prompt words required when employing the few-shot strategy. In practice, the same template as described in Sections C.1 and C.2 is utilized, thus it is not reiterated here.

\subsection{(Non-)Adversarial Prompt Template}

This template is designed to generate prompts that require a Large Language Model (LLM) to make judgments about the labels of sentences within it. Within the template, a simple example is used to aid the LLM in understanding the meanings of the three labels.

\begin{tcolorbox}[colback=gray!5!white, colframe=black, title=(Non-)Adversarial Prompt Template, breakable]
You are given a sentence that describes the content of a triplet in the form (Subject, Predicate, Object).
Each triplet and its corresponding sentence are classified into one of the three categories based on the accuracy of the content and the type of error:

\begin{enumerate}
    \item \textbf{true}: The triplet and the sentence accurately reflect the true content.
    \item \textbf{entity\_error}: The triplet contains an error related to the Subject or Object, which is reflected in the sentence.
    \item \textbf{predicate\_error}: The triplet contains an error related to the Predicate, which is reflected in the sentence.
\end{enumerate}

Here are some examples to help you understand:

\begin{enumerate}
    \item Sentence: "Albert Einstein discovered the Theory of Relativity." \\
          Category: true

    \item Sentence: "Albert Einstein discovered Radium." \\
          Category: entity\_error

    \item Sentence: "Albert Einstein disproved the Theory of Relativity." \\
          Category: predicate\_error
\end{enumerate}

Your task is to analyze the given sentence and determine which of the three categories it belongs to: "true," "entity\_error," or "predicate\_error."

Now, please classify the following sentence based on its content. Only output the category without anything else:

\textbf{Sentence:} "{\textcolor{red}{sentence}}" \\
\textbf{Category (choose one from true, entity\_error, predicate\_error):}
\end{tcolorbox}

\end{document}